\documentclass[10pt,twocolumn,letterpaper]{article}

\usepackage{iccv}
\usepackage{times}
\usepackage{epsfig}
\usepackage{graphicx}
\usepackage{amsmath}
\usepackage{amssymb}

\usepackage[pagebackref=true,breaklinks=true,letterpaper=true,colorlinks,bookmarks=false]{hyperref}

\iccvfinalcopy % *** Uncomment this line for the final submission

\def\iccvPaperID{3} % *** Enter the ICCV Paper ID here

\ificcvfinal\fi
\begin{document}

%%%%%%%%% TITLE
\title{ How Old Are You? \linebreak Face Age Translation with Identity Preservation Using GANs}

\author{
{Zipeng Wang, Zhaoxiang Liu\thanks{corresponding author}, Jianfeng Huang, Shiguo Lian, Yimin Lin}\\
Cloudminds\\
{\tt\small \{kohou.wang, robin.liu, jianfeng.huang, scott.lian, anson.lin\}@cloudminds.com}
% \and
% Zhaoxiang Liu\\
% Cloudminds Technologies Inc.\\
% % First line of institution2 address\\
% {\tt\small robin.liu@cloudminds.com}
% \and
% Shiguo Lian\\
% Cloudminds Technologies Inc.\\
% % First line of institution2 address\\
% {\tt\small scott.lian@cloudminds.com}
}

\maketitle
%\thispagestyle{empty}

%%%%%%%%% ABSTRACT
\begin{abstract}
   We present a novel framework to generate images of different age while preserving identity information, which is known as face aging. Different from most recent popular face aging networks utilizing Generative Adversarial Networks(GANs) application, our approach do not simply transfer a young face to an old one. Instead, we employ the edge map as intermediate representations, firstly edge maps of young faces are extracted, a CycleGAN-based network is adopted to transfer them into edge maps of old faces, then another pix2pixHD-based network is adopted to transfer the synthesized edge maps, concatenated with identity information, into old faces. In this way, our method can generate more realistic transfered images, simultaneously ensuring that face identity information be preserved well, and the apparent age of the generated image be accurately appropriate. Experimental results demonstrate that our method is feasible for face age translation.

\end{abstract}

%%%%%%%%% BODY TEXT
\section{Introduction}

Since the emergence of Generative Adversarial Network \cite{goodfellow2014generative}, a sequence of related works have been proposed \cite{zhu2017cycleGAN,isola2017pix2pix,
karras2017progressiveGAN,yu2017seqgan,
arjovsky2017wassersteinGAN, wang2018pix2pixHD, jin2017AnimeGAN, zhang2018SAGAN,chen2019realistic, wang2019neural, liu2019video}. Among which, pix2pix \cite{isola2017pix2pix} firstly takes conditional adversarial networks as a general-purpose solution to image-to-image translation issues, proposes a general framework for all such kind of issues, and their results verify that this method is quite effective at the task of translating one possible representation of a scene into another. This greatly promotes such kind of image-to-image translation tasks.

\begin{figure*}
    % \begin{figure}[t]
\begin{center}
% \fbox{\rule{0pt}{2in} \rule{.9\linewidth}{0pt}}
    \includegraphics[width=0.8\linewidth]{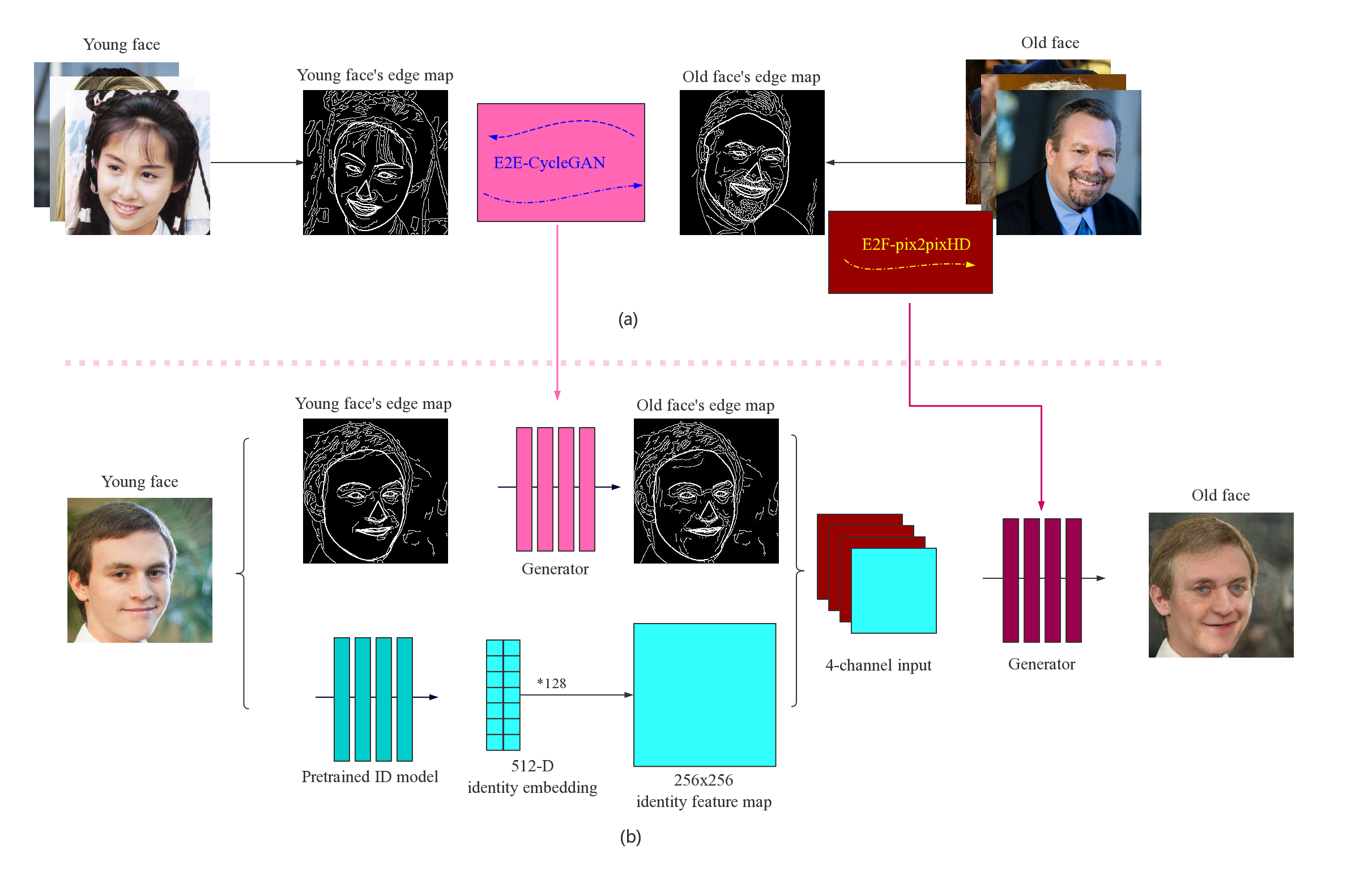}
\end{center}
   \caption{Our method. (a): Training procedure. Firstly edge maps of old face and young face are calculated, then these corresponding edge maps are utilized to train a E2E-CycleGAN model; meanwhile, edge maps and identity information of old face are utilized to train a E2F-pix2pixHD model, where original old face is the ground truth; (b): Inference procedure of translating young face into old face. Firstly, edge map of the young face will be calculated, then pretrained E2E-CycleGAN is utilized to generate corresponding old face's edge map, and finally pretrained E2F-pix2pixHD is utilized to generate the old face.}
\label{fig:our-method}
\end{figure*}
% \end{figure}

Face aging task, which means translating an image of a face into another image of a face to change its apparent age, has also received great attentions~\cite{Li2018GlobalAndLocalAgeGAN, Yang2017LearningFaceAge, zhao2018lookAcrossElapse, zhang2017ageGAN, palsson2018faceAgingGAN}. Nonetheless, earlier works mostly take face aging task as a simple image-to-image translation task, which simply trains GANs to translate a young face into an old one or vice versa. This kind of method leaves the network alone to learn the mapping between young faces and old ones, which still has room for improvement.

In this paper, we take one more step further to handle these problems. We investigate the face aging task as an image-to-image translation from edge maps to faces instead of directly translating young faces into old ones or vice versa. Our framework trains two GAN-based networks respectively: a CycleGAN-based network and a pix2pixHD-based network, which we call E2E-CycleGAN and E2F-pix2pixHD respectively, detailed descriptions are in Section~\ref{EEGAN} and Section~\ref{EFGAN}. The E2E-CycleGAN network is trained to translate young faces' edge maps into old ones, and also vice versa. Meanwhile we train another E2F-pix2pixHD network to translate faces' edge maps into faces. When transferring faces into different age, taking transferring young faces into old ones as example, the E2E-CycleGAN will be used to translate a young face's edge map into an old face's edge map, then the trained E2F-pix2pixHD generates old faces from the generated edge map, which is concatenated with identity information. Figure \ref{fig:our-method}(b) diagrams the procedure in detail. The reason why we use the edge map, which is the combination of canny contour and landmarks of a face image, is that we think the edge map is a crucial feature of a face image's apparent age. By utilizing edge map, we can focus the GANs' attention on the rather important texture information, simultaneously, utilizing texture information of background can also better describes an individual's properties, such as hairstyle, earrings, \etc. And also, because of the abundant texture information, our model can generate more photo-realistic and detailed facial expressions and emotions.

Our primary contribution is that: we propose a new framework to implement the age translation. Firstly we utilize E2E-CycleGAN to transfer face image's edge maps, then we utilize E2F-pix2pixHD to synthesize realistic faces with edge maps and identity information as the input. Our second contribution is that: utilizing this proposed framework, we can implement age translation of faces more realistically and delicately, while ensuring that face identity information be preserved well, and the apparent age of the generated image be more accurately appropriate.

%------------------------------------------------------------------------
\section{Related work}

{\bfseries Generative Adversarial Networks (GANs)} \cite{goodfellow2014generative} have achieved great success in content generation since proposed. This framework for estimating generative models via an adversarial process corresponds to a minimax two-player game. In the proposed \emph{adversarial nets} framework, the generative model is pitted against adversary: a discriminative model that learns to distinguish whether a sample is real or fake. Mathematically speaking, D and G play the following two-player minimax game with value function $V(G,D)$:
% \begin{equation}
% \begin{split}
\begin{align*}
\underset{G}{min} \underset{D}{max} V(D,G)&=\mathbb{E}_{x\sim P_{data}(x))}[logD(x)]\\
 &+\mathbb{E}_{z\sim P_{z}(z))}[log(1-D(G(z)]
\end{align*}
\label{equation-GAN}
% \end{equation}
In the case where G and D are defined by multilayer perceptrons, the entire Generative Adversarial Network can be trained with backpropagation. There is no need for any Markov chains or unrolled approximate inference networks during either training or generation of samples, which greatly simplifies the overall procedure.

\begin{figure}[t]
\begin{center}
   \includegraphics[width=0.8\linewidth]{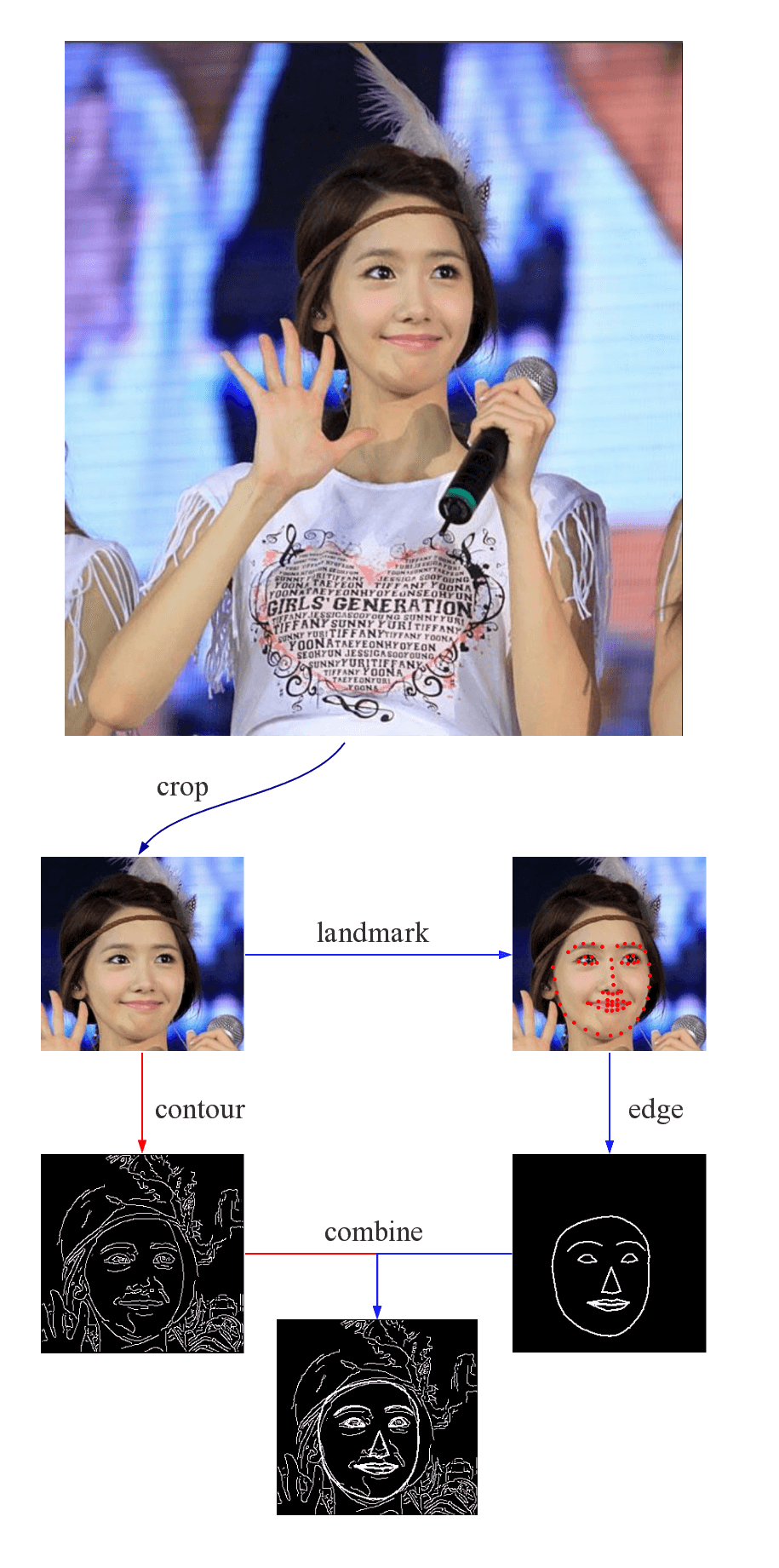}
\end{center}
   \caption{Face ROI workflow. Firstly a face ROI is cropped utilizing a pretrained face detector model, then landmarks is detected and thereafter the edge is drawn; and the canny contour is calculated, which is joined together with edge to be the final input edge map.}
\label{fig:ROI-workflow}
% \label{fig:onecol}
\end{figure}

 \begin{figure}[t]
 \begin{center}
    \includegraphics[width=0.8\linewidth]{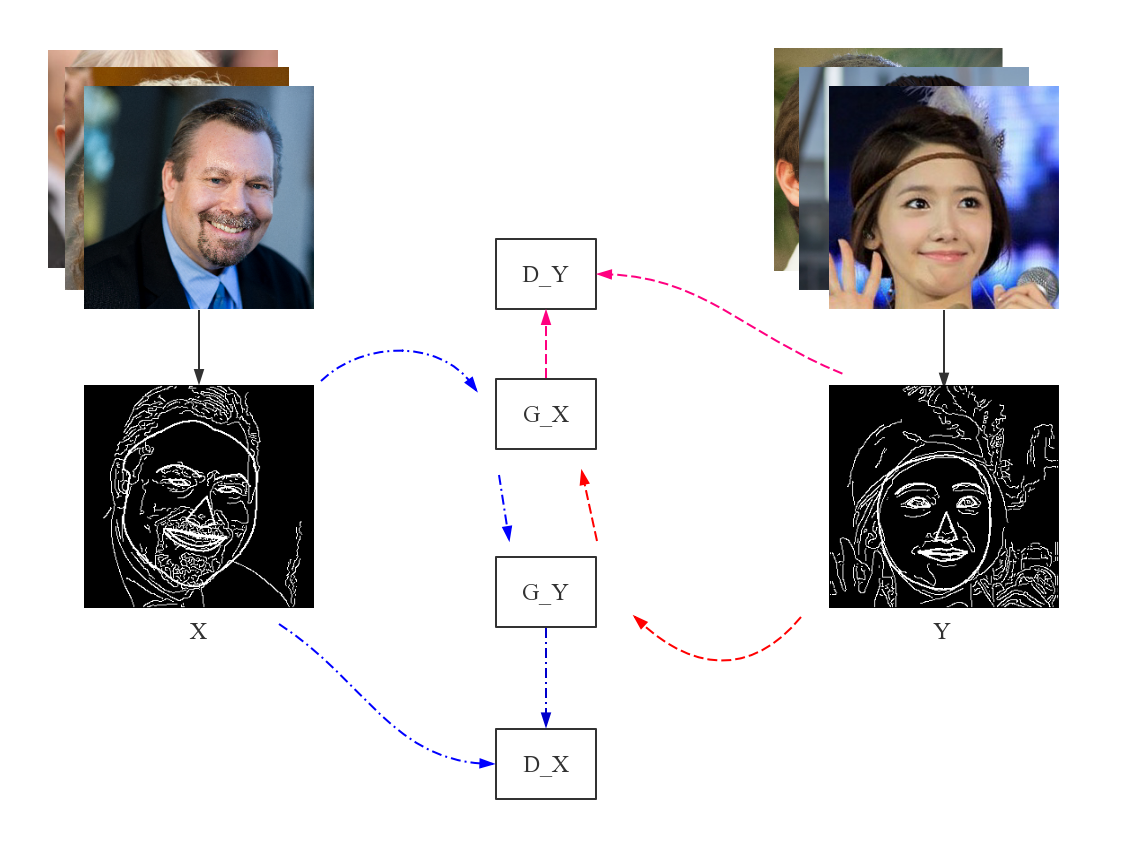}
 \end{center}
    \caption{E2E-CycleGAN training procedure. Firstly edge maps of young face images and old face images are calculated, then these two kinds of edge maps will serve as two different domains to perform the E2E-CycleGAN training. $D_X$ and $G_X$ denote the discriminator and  generator network of image set X, while $D_Y$ and $G_Y$ denote the discriminator and generator of image set Y.}
 \label{fig:CycleGAN_training}
 % \label{fig:onecol}
 \end{figure}

{\bfseries Image-to-Image Translation} Benefited from the arise of GANs \cite{goodfellow2014generative}, Isola \etal \cite{isola2017pix2pix} present an image-to-image framework pix2pix utilizing conditional adversarial network \cite{mirza2014conditionalGAN}. They investigate conditional adversarial networks as a general-purpose solution to image-to-image translation problems, which learns the mapping from input images to output images while simultaneously learning a loss function to train this mapping. By several experiments, they demonstrate that on a wide variety of problems, conditional GANs produce reasonable results, and present a simple framework sufficient enough to achieve good results and to analyze the effects of several important architectural choices. Youssef \etal~\cite{mejjati2018attentionGAN} propose unsupervised attention mechanisms that are jointly adversarially trained with the generators and discriminators to synthesize more realistic mappings. Chen and Hays~\cite{chen2018sketchygan} introduce a SketchyGAN to synthesize realistic images from human drawn sketches. Regmi and Borji \cite{regmi2018cross-view} tackle the problem of synthesizing ground-level images from overhead imagery and vice versa. Wang~\etal~\cite{wang2018pix2pixHD} further present a method utilizing conditional GANs, which can generate $2048*1024$ photo-realistic results. They utilize a novel adversarial loss and new multi-scale generator and discriminator architectures. Our approach builds on this pix2pixHD framework to learn the mapping from edge maps to real faces, Detailed description is in Section~\ref{EFGAN}.

\begin{figure*}[t]
\begin{center}
   \includegraphics[width=0.8\linewidth]{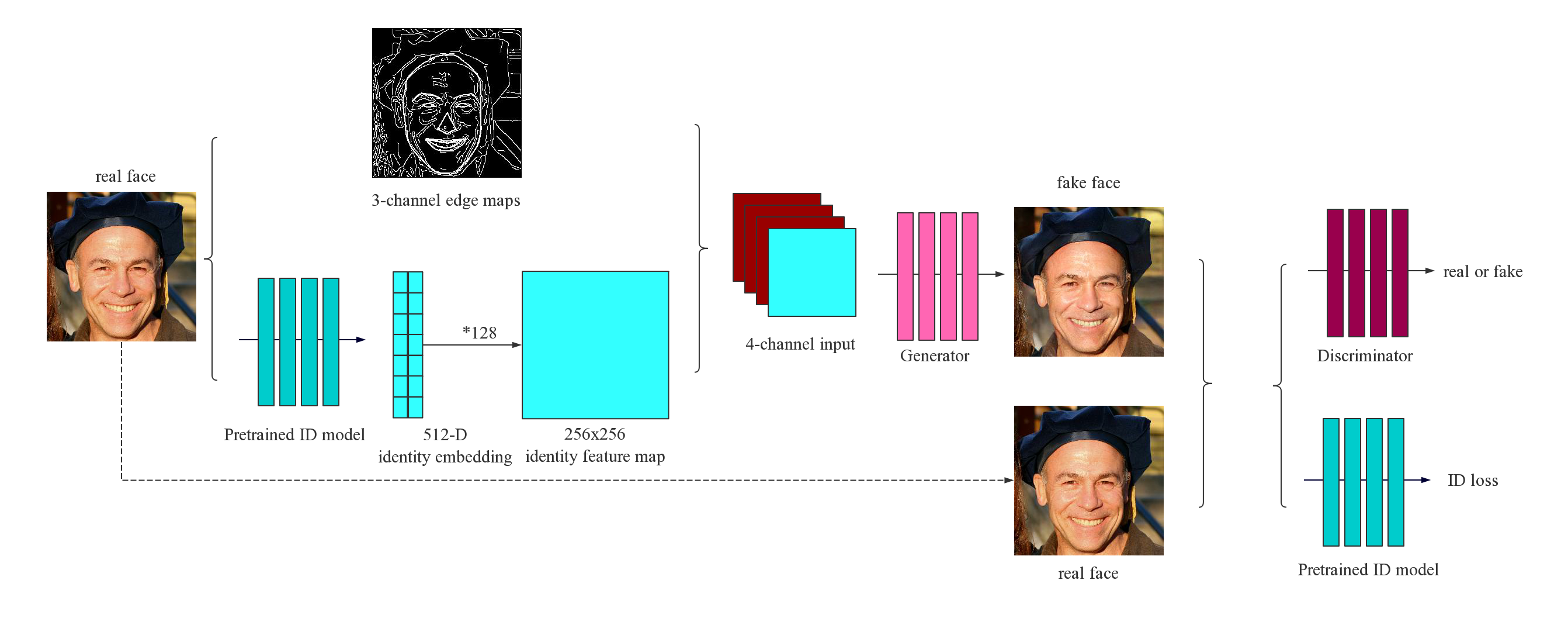}
\end{center}
   \caption{E2F-pix2pixHD training procedure. Firstly an edge map will be calculated using face landmarks and canny contour, then 512-D identity embedding will be calculated using a pretrained identity model, it will then be duplicated 128 times to construct a $256*256$ feature map; afterwards edge map and the feature map of identity embedding are concatenated together to be the generator's input. A discriminator will be trained to distinguish fake images from real ones.}
\label{fig:pix2pix_training}
% \label{fig:onecol}
\end{figure*}

\begin{figure}%[t]
\begin{center}
   \includegraphics[width=0.8\linewidth]{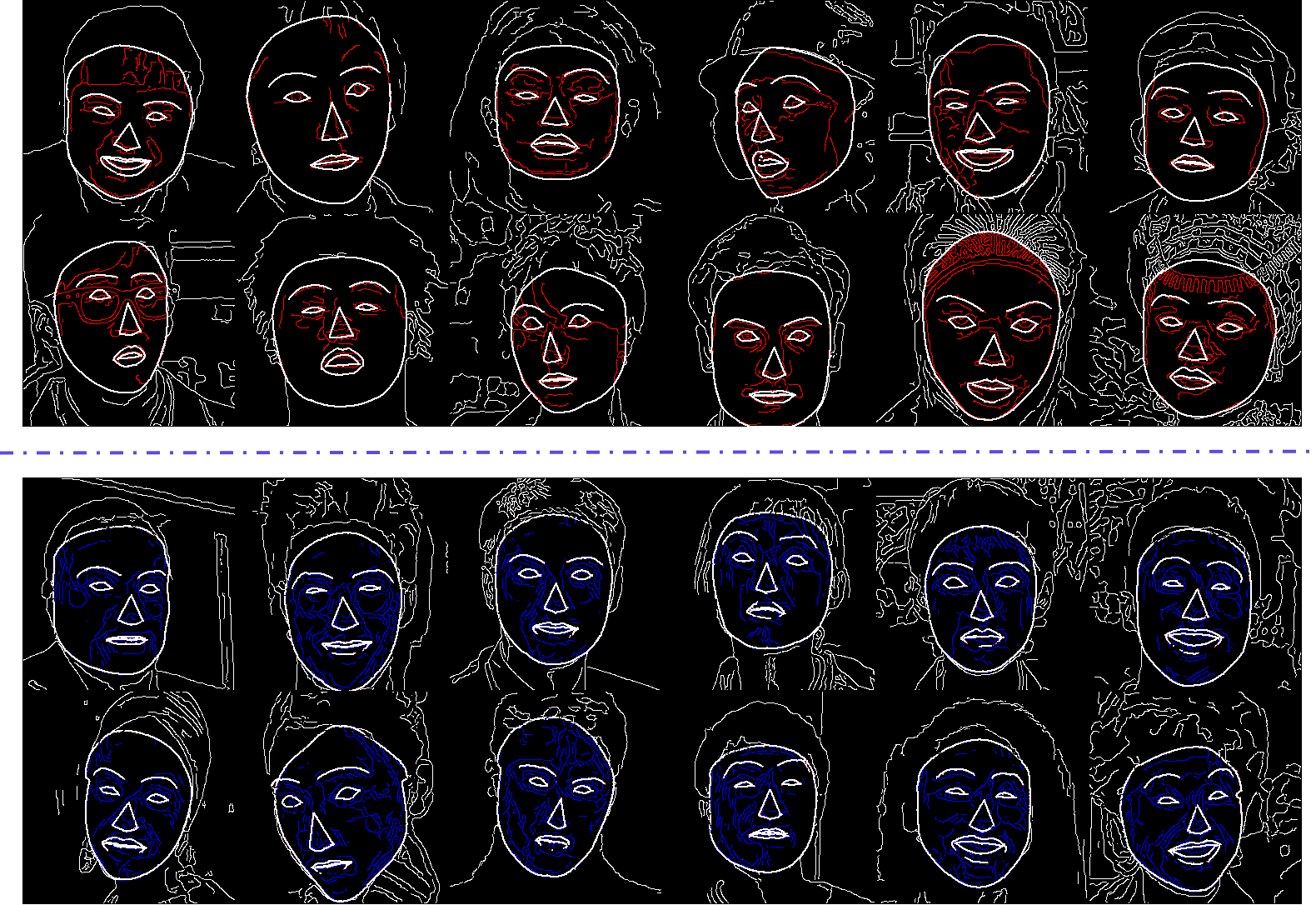}
\end{center}
   \caption{Edge maps for E2E-CycleGAN model. The upper half are the edge maps of the young face images, whose canny contour inside the face ROI's edge is set red; while the lower half are the edge maps of the old ones, whose canny contour inside the face ROI's edge is set green. In this way we can focus the model's attention on the changes of facial texture.}
\label{fig:red_blue}
% \label{fig:onecol}
\end{figure}

{\bfseries Unpaired Image-to-Image Translation} Since thousands of paired training data may not always be available, the unpaired setting also received enthusiastic attentions. Yi \etal~\cite{yi2017dualGAN} present a dual learning framework DualGAN, which simultaneously learns two reliable image translators from one domain to the other and hence can operate on a wide variety of image-to-image translation tasks. Zhu \etal~\cite{zhu2017cycleGAN} present a cycle-consistent adversarial network CycleGAN, which learns the mapping between two image collections by trying to capture correspondences between higher-level appearance structures. In this way, this method can also be applied to other tasks, such as painting to photo, and object transfiguration. Our method also utilizes this CycleGAN framework to learn the mapping from edge maps to edge maps, detailed description is in Section~\ref{EEGAN}.

{\bfseries Face Aging} Recent approaches with GANs also tackle the face aging problem. Zhang \etal~\cite{zhang2017ageGAN} present a GAN-based approach, which separately models the constraints for the intrinsic subject-specific characteristics and the age-specific facial changes with respect to the elapsed time, so as to ensure that the generated faces present desired aging effects while simultaneously keeping personalized properties stable. Palsson \etal~\cite{palsson2018faceAgingGAN} also consider the age of the person as the underlying style of the image, and present F-GAN, which is a fusion of two separately trained GANs. Li~\etal~\cite{Li2018GlobalAndLocalAgeGAN} propose a Global and Local Consistent Age Generative Adversarial Network(GLCA-GAN) to synthesize better results. They construct a generator which contains a global network and three local networks to learn the whole facial structure and imitate subtle changes of crucial facial subregions simultaneously. The input to local networks are three subregions of eyes, snouts and forehead from the whole face respectively, which is meant to imitate texture changes. Yang~\etal ~\cite{Yang2017LearningFaceAge} present an approach which models the constraints for the intrinsic subject-specific characteristics and also the age-specific facial changes with respect to the elapsed time respectively. They further utilize a pyramidal adversarial discriminator for high-level age-specific features' consistency. Zhao~\etal~\cite{zhao2018lookAcrossElapse} propose a novel Age-Invariant Model(AIM) for joint disentangled representation learning and photorealistic cross-age face synthesis, which learns to generate age-invariant facial representations explicitly disentangled from the age variation. Different from the above approaches, our focus is the edge map of faces, and the training procedure is shown in Figure \ref{fig:our-method}(a). By utilizing the edge map, our method can preserve the important texture information, and synthesize more realistic images whose apparent ages are more appropriate and facial details more abundant. The experiments results will be given to show our framework's effectiveness, detailed description is in Section~\ref{implementation}.

\begin{figure}[t]
\begin{center}
   \includegraphics[width=0.8\linewidth]{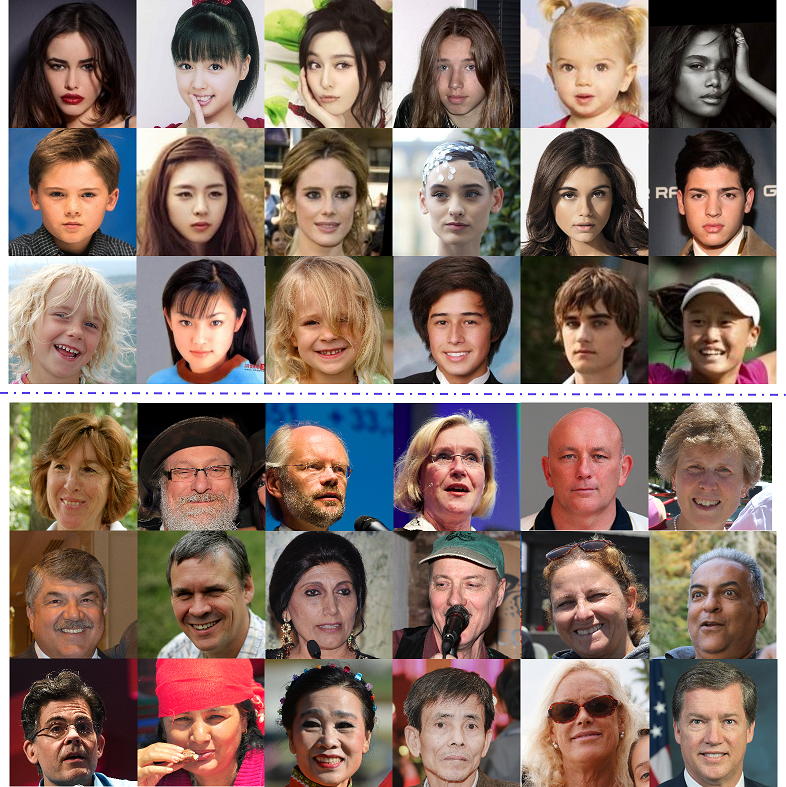}
\end{center}
   \caption{Training dataset for E2E-CycleGAN. The upper half is the young faces, while lower half is old faces. Our dataset has a good coverage of ethnicity, occlusions, poses, \etc.}
\label{fig:glimpse_of_dataset.}
% \label{fig:onecol}
\end{figure}

%-------------------------------------------------------------------------
\section{Proposed methods}\label{Whole_method}

Conditional GANs are generative models that learn a mapping from observed image $x$ and random noise $z$ to output image $y$, $G:{x,z}\to y$ \cite{mirza2014conditionalGAN}, while both the generator and discriminator are conditioned on the extra information $x$. In this paper, the proposed framework mainly builds on this conditional GANs.

As is diagrammed in Figure~\ref{fig:our-method}, we lay emphasis on the edge maps of face images, which is the combination of canny contour and landmarks of a face image. We think edge map of face image as crucial feature of a face image's apparent age, the details we compute the edge map of an image are in Section~\ref{preprocessing}. In this way, the edge map can preserve most important information of face's apparent age, while utilizing texture information of background can also better describes an individual's facial properties. And also, because of the abundant texture information, our model can generate more photo-realistic and detailed facial expressions and emotions.

\begin{figure}[t]
\begin{center}
   \includegraphics[width=0.8\linewidth]{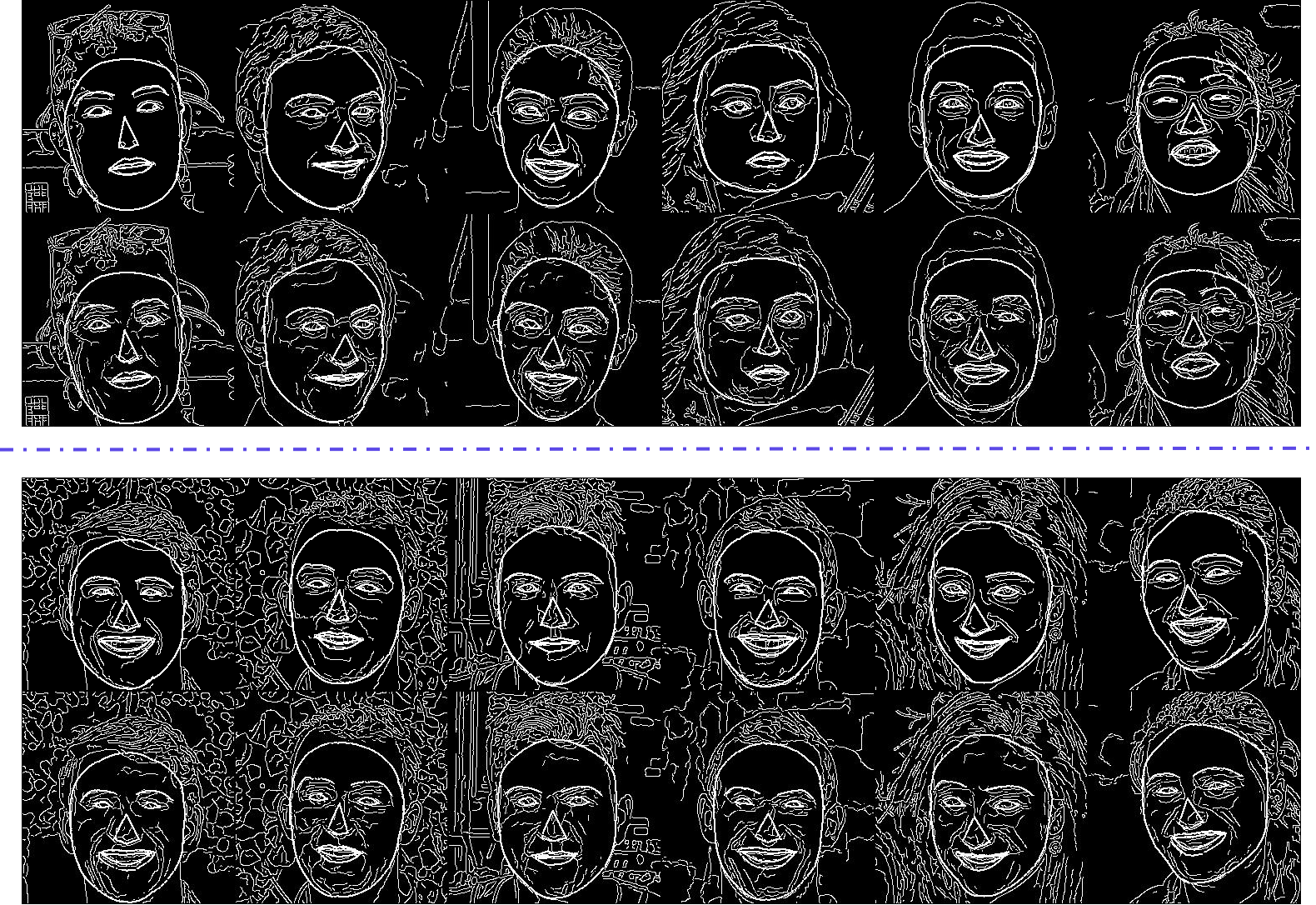}
\end{center}
   \caption{Synthesized edge maps of E2E-CycleGAN. The first row is the input young edge maps, second row is synthesized old edge maps, while third row input young edge maps, and fourth row synthesized old edge maps. As can be seen from pictures, our method can preserve the overall outline well while still adding appropriate texture information to the edge maps.}
\label{fig:CycleGAN_results}
% \label{fig:onecol}
\end{figure}

\begin{figure}%[t]
\begin{center}
   \includegraphics[width=0.8\linewidth]{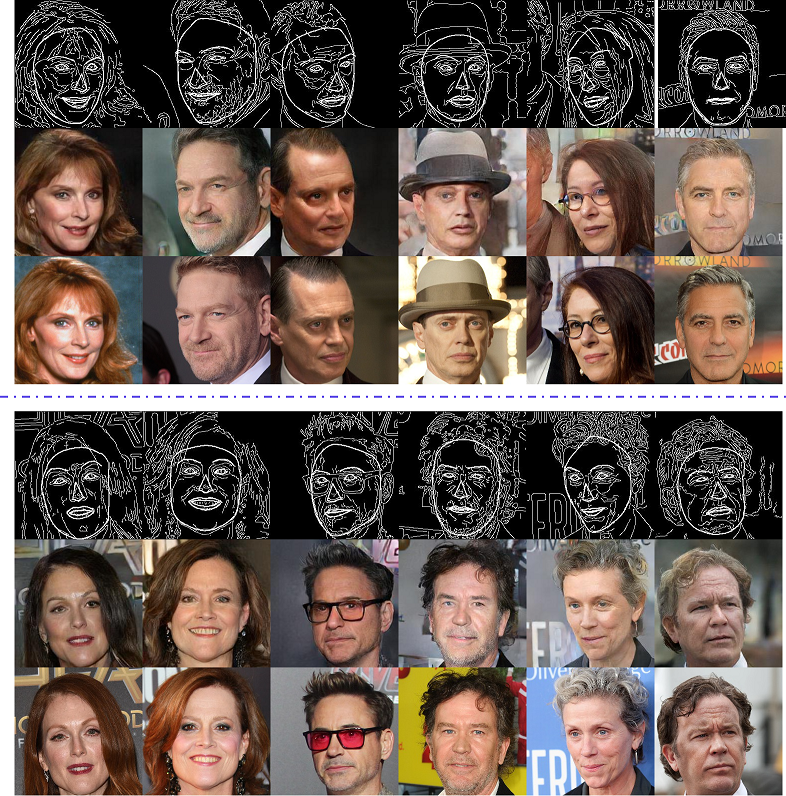}
\end{center}
   \caption{Synthesized faces of E2F-pix2pixHD. The first row and the fourth row are the edge maps, second row and fifth row are synthesized faces, and third row and sixth row are ground truth faces. As can be seen from the pictures, our method can preserve identity information well and the apparent age of synthesized faces are close to the ground truth's.}
\label{fig:EFGAN_results}
% \label{fig:onecol}
\end{figure}
%-------------------------------------------------------------------------
\subsection{Preprocessing}\label{preprocessing}
 Images will firstly be cropped utilizing a pretrained face detector model RSA~\cite{liu2017RSA}. To include more information of hairstyle, ears, hats, \etc, we crop a 1.5 times bigger ROI where the face ROI is in the center instead of the face ROI only. Then a landmark detector is adopted to detect the 68 landmarks of face, afterwards the 68 landmarks are connected together to draw the contour of faces. This contour will then be joined together with the canny contour of the face, combining into the final edge map. this process is diagrammed in Figure \ref{fig:ROI-workflow}.

%-------------------------------------------------------------------------
\subsection{E2E-CycleGAN: Edge to edge with CycleGAN}\label{EEGAN}
As discussed before, CycleGAN, presented by Zhu \etal~\cite{zhu2017cycleGAN}, learns the mappings from two image collections, and does not require the dataset to be paired images. CycleGAN introduces a cycle consistency loss so as to learn the mappings without paired training examples. Firstly the adversarial loss \cite{goodfellow2014generative} is adopted, the objective for the mapping function $G:X \to Y$ and its discriminator $D_Y$ is:
% \begin{equation}
% \begin{split}
\begin{align*}
    L_{GAN}(G,D_Y,X,Y)&=\mathbb{E}_{y\sim P_{data}(y)}[logD_Y(y))]\\
     &+\mathbb{E}_{x\sim P_{data}(x))}[log(1-D_Y(G(x))]
\end{align*}
\label{equation-CycleGAN_loss1}
% \end{equation}
$G$ tries to generate images $G(x)$ that can cheat D from distinguishing $G(x)$ from real images $y$. And also, for the adverse mapping function $F:Y \to X$ :
% \begin{equation}
% \begin{split}
\begin{align*}
    L_{GAN}(F,D_X,Y,X)&=\mathbb{E}_{x\sim P_{data}(x)}[logD_X(x))]\\
     &+\mathbb{E}_{y\sim P_{data}(y)}[log(1-D_X(F(y))]
\end{align*}
\label{equation-CycleGAN_loss2}
% \end{equation}

To further reduce the space of possible mapping functions, as adversarial loss surely can not guarantee an input image will be translated into an image of target domain, Zhu \etal~\cite{zhu2017cycleGAN} introduce the cycle consistent loss:
% \begin{split}
% \usepackage{amsmath}
% \begin{equation}
\begin{align*}
    L_{cyc}(G,F)&=\mathbb{E}_{x\sim P_{data}(x)}[||F(G(x))-x||_1]\\
     & +\mathbb{E}_{y\sim P_{data}(y)}[||G(F(y))-y||_1]
\end{align*}
\label{equation-CycleGAN_cycleLoss}
% \end{equation}
This is inspired by the \emph{forward cycle consistency}: an input image x should be similarly the same after translated to y and then translated back, \ie, $X \to G(x) \to F(G(x)) \approx x$. The full objective is:
% \begin{equation}
\begin{align*}
     L(G,F,D_X,D_Y)&= L_{GAN}(G,D_Y,X,Y)\\&+ L_{GAN}(F,D_X,Y,X)\\&+  \lambda L_{cyc}(G,F)
\end{align*}
\label{equation-CycleGAN_FUllLoss}
% \end{equation}

The whole training procedure is detailed in Figure \ref{fig:CycleGAN_training}. In this paper, the input and output for CycleGAN training procedure are the edge maps of old faces and young faces respectively, which is why we call the network E2E-CycleGAN. The edge maps are computed as the procedure described in Section~\ref{preprocessing}.

To better focus the model's attention on the canny contour inside face ROI's edge, when training E2E-CycleGAN model, we set the young face images' canny contour inside face ROI's edge to red color, and old face image's canny contour to green color, the rest canny contour and edge are white as before. An illustration of this method is as Figure~\ref{fig:red_blue} shows. In this way, we manually focus the model's attention on the canny contour inside the face ROI's edge, which represents the changes of facial texture and is an important feature of different ages. We will set the color of canny contour back to white after the synthesis of edge maps, which serves as the input of E2F-pix2pixHD model.

%-------------------------------------------------------------------------
\subsection{E2F-pix2pixHD: Edge to face with pix2pixHD}\label{EFGAN}
Isola \etal~\cite{isola2017pix2pix} present a conditional GAN framework pix2pix, which prevails on many image-to-image tasks, but the results are rather limited to low-resolution. And Wang~\etal~\cite{wang2018pix2pixHD} present another GAN-based framework pix2pixHD, which can synthesize $2048*1024$ photo realistic results. We utilize this pix2pixHD framework on our task of face aging and call this network E2F-pix2pixHD. In our proposed framework, firstly, a face landmark detector is adopted to detect the 68 landmarks of faces, these 68 landmarks will then be connected to draw a face's contour; Secondly a canny operator is adopted to edge filter an image using canny algorithm. Then the face edge and canny edge will be joined together into an image, which is a part of the input for our E2F-pix2pixHD model. The other part is the identity embedding. We use a pretrained face model ArcFace~\cite{deng2018arcface} to extract the face ROI's 512-D identity feature embedding, then duplicate it 128 times to construct a $256*256$ identity feature map. Then we concatenate the face ROI's identity feature map with original edge map, hence get a 4 channel input, while the output is the face image itself. This procedure is diagrammed in Figure \ref{fig:pix2pix_training}.

In this way, the additional identity embedding can better direct the generator to generate more realistic images while preserving sufficient identity information. We consider the edge map and identity information are sufficient enough for the generation of a human face, while still preserving the identity properties properly, such as hairstyle, earrings, \etc. Section \ref{section:pix2pix-results} details the results that demonstrate this standpoint.

%-------------------------------------------------------------------------
\subsection{Generation of faces}
Our generation procedure is detailed in Figure \ref{fig:our-method}(b), taking translating young faces into old faces as an example, firstly an edge map $E_x$ of young face $x$ will be drawn, then the trained $G_{E2E-CycleGAN}$ will use this $E_x$ as an input to generate an corresponding old face's edge map $E_y$, the $E_y$, concatenated with identity information, will then be used to generate an old face $y$ by trained $G_{E2F-pix2pixHD}$. The results are detailed in Section \ref{section:full_frame-results}.

\begin{figure}[t]
\begin{center}
   \includegraphics[width=0.8\linewidth]{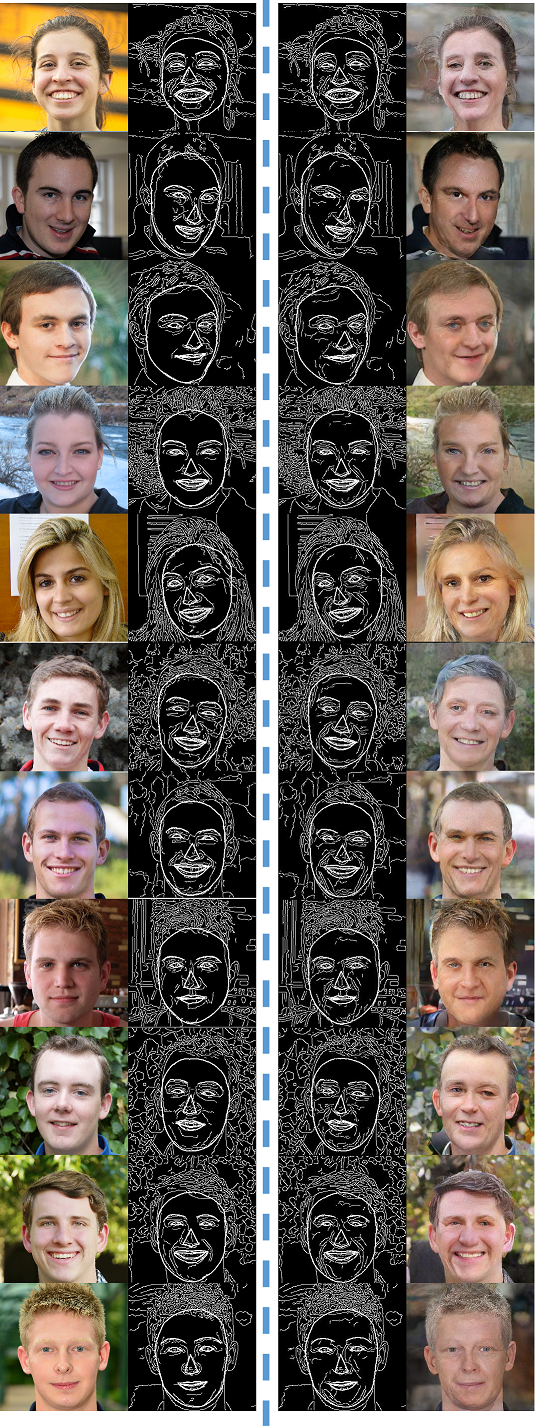}
\end{center}
   \caption{Synthesized faces of whole model. The first column is the young face images, second column is the young edge maps, third column is the synthesized old edge maps, while fourth column is synthesized fake old images.}
\label{fig:whole_results}
% \label{fig:onecol}
\end{figure}

\begin{figure}%[t]
\begin{center}
   \includegraphics[width=0.8\linewidth]{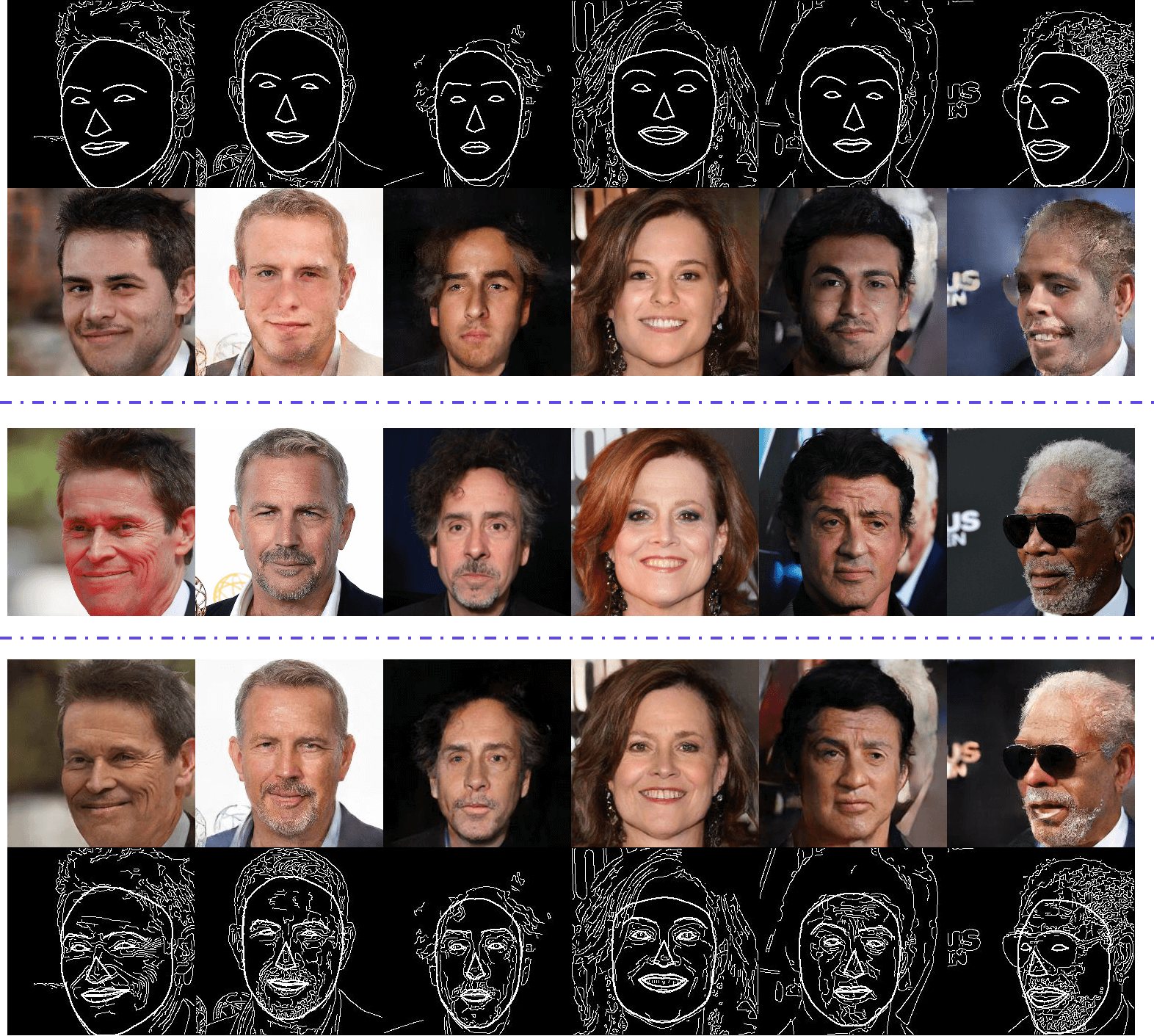}
\end{center}
   \caption{Different results of two E2F-pix2pixHD models. The first row and the last row are the two models' input edge maps, second row and fourth row are their synthesized results according to the input, and third row is ground truth faces. As can be seen from the results, the canny contour inside face ROI's edge serves as an important feature for the consistency of the ages.}
\label{fig:ablation_results}
% \label{fig:onecol}
\end{figure}

%-------------------------------------------------------------------------
\section{Implementation}\label{implementation}

%-------------------------------------------------------------------------
\subsection{Datasets}\label{datasets}
Our datasets are obtained from CelebA~\cite{datasets-CelebA} and FFHQ~\cite{karras2018styleGAN}. CelebA~\cite{datasets-CelebA}(CelebFaces Attributes Dataset) is a large-scale face attributes dataset with more than 200K celebrity images. The images in this dataset cover large pose variations and background clutter, including 10,177 number of identities, in total 202,599 number of face images. FFHQ~\cite{karras2018styleGAN}(Flickr-Faces-HQ) is another high-quality image dataset of human faces, consisting of 70,000 high-quality images at $1024*1024$ resolution and containing considerable variation in terms of age, ethnicity and image background. It also has good coverage of accessories such as makeups, occlusions, etc. The images were crawled from Flickr, thus inheriting all the biases of that website.

In this paper, we firstly process CelebA and FFHQ datasets as training set. After preprocessing, we further utilize a pretrained age model to detect the face ROI's age, so as to classify images into age groups, such as old group and young group, where old means age is no lower than 60, and young means age no higher than 28. For the E2E-CycleGAN training, we get 6k young images and 28k old images, a glimpse of the training dataset is as Figure~\ref{fig:glimpse_of_dataset.}. As can be seen from the pictures, our training dataset covers a good deal of variations, such as expressions, ethnicity, backgrounds, \etc. And for E2F-pix2pixHD training, we pick out 51k images of different ages so as to ensure the E2F-pix2pixHD model can generalize well to all kinds of faces, especially for the synthesis of various ages.

% In this paper, we firstly process CelebA and FFHQ datasets as training set. After preprocessing, we further utilize a pretrained age model to detect the face ROI's age, so as to classify images into age groups, such as old group and young group, where old means age is no lower than 60, and young means age no higher than 28. For the E2E-CycleGAN training, we get 6k young images and 28k old images; and for E2F-pix2pixHD training, we pick out 51k images of different ages so as to ensure the E2F-pix2pixHD model can generalize well to all kinds of faces, especially for the synthesis of various ages. A glimpse of the training dataset is as Figure~\ref{fig:glimpse_of_dataset.}. As can be seen from the pictures, our training dataset covers a good deal of variations, such as expressions, ethnicity, backgrounds, \etc.

%-------------------------------------------------------------------------
\subsection{Training details of E2E-CycleGAN}\label{section:CycleGAN-results}
For E2E-CycleGAN training, we firstly preprocess dataset to calculate edge maps, then these edge maps are utilized to train the model, in this section we will elaborate this procedure.

Firstly two groups of face images are selected according to their detected ages. The face in one group are younger than 28 while the other group are older than 60, the young group has more than 6k images while old group has more than 28k images. Then we calculate the edge maps of the two group face images, which will be the two groups of training set. In this training, we set a minimum loss at 1.0. After training, we get two generators, one is young-to-old while the other is old-to-young. The young-to-old model is fed with face edge maps younger than 28 and outputs face edge maps older than 60, while old-to-young model transfers old edge maps into edge maps younger than 28.

%-------------------------------------------------------------------------
\subsection{Training details of E2F-pix2pixHD }\label{section:pix2pix-results}
For E2F-pix2pixHD training, we clean out more than 51k face images of different ages, we train the E2F-pix2pixHD model on these images, so as to ensure the synthesized images can generalize to all kinds of faces during different periods of ages, including glasses, ethnicity, expressions, \etc. All of these training procedure utilize face images' edge maps and identity embeddings as input, as is described in Section~\ref{EFGAN}.

%-------------------------------------------------------------------------
\subsection{Evaluation of whole model}\label{section:full_frame-results}
In this section we will illustrate the results of out models orderly. The synthesized edge maps for translating young edge maps to old edge maps, which is done by E2E-CycleGAN model, are shown as Figure~\ref{fig:CycleGAN_results}. As can be seen from pictures, our method can preserve the overall outline of face well while still adding appropriate texture information to the edge maps, which is crucial for the synthesis of an older face image.

The synthesized images of E2F-pix2pixHD model are shown as Figure~\ref{fig:EFGAN_results}. As can be seen from the pictures, our method can preserve identity well, meanwhile the apparent age of synthesized image is also much appropriate.

As shown in Figure~\ref{fig:our-method}(b), our synthesis procedure of a young face to old face mainly depends on the edge map. Firstly, an edge map of a young face will be calculated, which will be utilized by the pretrained E2E-CycleGAN to synthesize an old edge map. Afterwards the pretrained E2F-pix2pixHD will use this old edge map, along with identity information, to synthesize an image of old face. In this way, we hold the view that the model can synthesize more realistic face images, whose apparent ages are more appropriate, while still preserving identity information well. Figure~\ref{fig:whole_results} shows the synthesized images of our whole framework. As can be seen, the results are much photorealistic, and represent the aging of faces well. And also, the texture added by E2E-CycleGAN serves as a good feature for the synthesis of older faces.

%-------------------------------------------------------------------------
\subsection{Ablation study}\label{section:ablation-study}
To better explore our method's efficiency and effectiveness, we further conduct an ablation study to demonstrate the edge map's effect.

In details, we train two E2F-pix2pixHD models separately: one with edge maps of faces as input, but the canny contour inside the face ROI's edge is filtered out, and the other model's input is edge map as we described in Section~\ref{preprocessing}. We train the two models with same datasets, which is the 51k merged images we cleaned out in Section~\ref{datasets}, and train them for same 200 epochs. Finally we test the two models with same test data, which are several random images of IMDB-WIKI~\cite{IMDB-wikiData} dataset. In this way we control all other conditions the same except the canny contour inside the face ROI's edge, so as to ensure that the two models' performance will have difference only because of this difference.

And we measure the two model's performance through measuring the generated images. The results are shown in Figure~\ref{fig:ablation_results}. With canny contour inside the face ROI's edge, the model can generate more realistic and consistent images of the ground truth, the synthesized images have more abundant facial details and the age is more accurate with the ground truth. Moreover, the model without canny contour inside the face ROI's edge can't well generate the details of faces, and even worse, it seems to casually add or subtract glasses of faces, which is a big drawback of the synthesis of face images. Most importantly, without the canny contour inside face ROI's edge, the model can't well infer the input edge map's age, which leads to the results that the synthesized images seems to have all kinds of casual apparent ages. On the contrary, our method can well generate images of more correct apparent ages, which is benefited from the added texture information.

%-------------------------------------------------------------------------
\section{Conclusion}
In this paper, we propose a new framework to synthesize images of faces with identity information well preserved and apparent age much appropriate, and the experimental results demonstrate that our method is feasible for face age translation. Although our method shows compelling results, it sometimes may also synthesize results not so satisfactory, such as ears of the same people is not exactly mirrored symmetrically, also hair may have abnormal distribution, \etc. Nonetheless, in most cases our method can synthesize a much realistic face image while preserving identity well and the apparent age is much more appropriate.

% \begin{table}
% \begin{center}
% \begin{tabular}{|l|c|}
% \hline
% Method & Frobnability \\
% \hline\hline
% Theirs & Frumpy \\
% Yours & Frobbly \\
% Ours & Makes one's heart Frob\\
% \hline
% \end{tabular}
% \end{center}
% \caption{Results.   Ours is better.}
% \end{table}

%\newpage
%\pagebreak

%\newpage
%\pagebreak

{\small
\bibliographystyle{ieee}
\bibliography{egbib}

\begin{thebibliography}{10}\itemsep=-1pt

\bibitem{arjovsky2017wassersteinGAN}
M.~Arjovsky, S.~Chintala, and L.~Bottou.
\newblock Wasserstein gan.
\newblock {\em arXiv preprint arXiv:1701.07875}, 2017.

\bibitem{chen2018sketchygan}
W.~Chen and J.~Hays.
\newblock Sketchygan: towards diverse and realistic sketch to image synthesis.
\newblock In {\em Proceedings of the IEEE Conference on Computer Vision and
  Pattern Recognition}, pages 9416--9425, 2018.

\bibitem{chen2019realistic}
Z.~Chen, Z.~Liu, H.~Hu, J.~Bai, S.~Lian, F.~Shi, and K.~Wang.
\newblock A realistic face-to-face conversation system based on deep neural
  networks.
\newblock {\em arXiv preprint arXiv:1908.07750}, 2019.

\bibitem{deng2018arcface}
J.~Deng, J.~Guo, X.~Niannan, and S.~Zafeiriou.
\newblock Arcface: Additive angular margin loss for deep face recognition.
\newblock In {\em CVPR}, 2019.

\bibitem{goodfellow2014generative}
I.~Goodfellow, J.~Pouget-Abadie, M.~Mirza, B.~Xu, D.~Warde-Farley, S.~Ozair,
  A.~Courville, and Y.~Bengio.
\newblock Generative adversarial nets.
\newblock In {\em Advances in neural information processing systems}, pages
  2672--2680, 2014.

\bibitem{isola2017pix2pix}
P.~Isola, J.-Y. Zhu, T.~Zhou, and A.~A. Efros.
\newblock Image-to-image translation with conditional adversarial networks.
\newblock In {\em Proceedings of the IEEE conference on computer vision and
  pattern recognition}, pages 1125--1134, 2017.

\bibitem{jin2017AnimeGAN}
Y.~Jin, J.~Zhang, M.~Li, Y.~Tian, H.~Zhu, and Z.~Fang.
\newblock Towards the automatic anime characters creation with generative
  adversarial networks.
\newblock {\em arXiv preprint arXiv:1708.05509}, 2017.

\bibitem{karras2017progressiveGAN}
T.~Karras, T.~Aila, S.~Laine, and J.~Lehtinen.
\newblock Progressive growing of gans for improved quality, stability, and
  variation.
\newblock {\em arXiv preprint arXiv:1710.10196}, 2017.

\bibitem{karras2018styleGAN}
T.~Karras, S.~Laine, and T.~Aila.
\newblock A style-based generator architecture for generative adversarial
  networks.
\newblock {\em arXiv preprint arXiv:1812.04948}, 2018.

\bibitem{Li2018GlobalAndLocalAgeGAN}
P.~Li, Y.~Hu, L.~Qi, H.~Ran, and Z.~Sun.
\newblock Global and local consistent age generative adversarial networks.
\newblock 2018.

\bibitem{liu2017RSA}
Y.~Liu, H.~Li, J.~Yan, F.~Wei, X.~Wang, and X.~Tang.
\newblock Recurrent scale approximation for object detection in cnn.
\newblock In {\em Proceedings of the IEEE International Conference on Computer
  Vision}, pages 571--579, 2017.

\bibitem{liu2019video}
Z.~Liu, H.~Hu, Z.~Wang, K.~Wang, J.~Bai, and S.~Lian.
\newblock Video synthesis of human upper body with realistic face.
\newblock {\em arXiv preprint arXiv:1908.06607}, 2019.

\bibitem{datasets-CelebA}
Z.~Liu, P.~Luo, X.~Wang, and X.~Tang.
\newblock Deep learning face attributes in the wild.
\newblock In {\em Proceedings of International Conference on Computer Vision
  (ICCV)}, 2015.

\bibitem{mejjati2018attentionGAN}
Y.~A. Mejjati, C.~Richardt, J.~Tompkin, D.~Cosker, and K.~I. Kim.
\newblock Unsupervised attention-guided image-to-image translation.
\newblock In {\em Advances in Neural Information Processing Systems}, pages
  3697--3707, 2018.

\bibitem{mirza2014conditionalGAN}
M.~Mirza and S.~Osindero.
\newblock Conditional generative adversarial nets.
\newblock {\em arXiv preprint arXiv:1411.1784}, 2014.

\bibitem{palsson2018faceAgingGAN}
S.~Palsson, E.~Agustsson, R.~Timofte, and L.~Van~Gool.
\newblock Generative adversarial style transfer networks for face aging.
\newblock In {\em Proceedings of the IEEE Conference on Computer Vision and
  Pattern Recognition Workshops}, pages 2084--2092, 2018.

\bibitem{regmi2018cross-view}
K.~Regmi and A.~Borji.
\newblock Cross-view image synthesis using conditional gans.
\newblock In {\em Proceedings of the IEEE Conference on Computer Vision and
  Pattern Recognition}, pages 3501--3510, 2018.

\bibitem{IMDB-wikiData}
R.~Rothe, R.~Timofte, and L.~V. Gool.
\newblock Deep expectation of real and apparent age from a single image without
  facial landmarks.
\newblock {\em International Journal of Computer Vision (IJCV)}, July 2016.

\bibitem{wang2018pix2pixHD}
T.~C. Wang, M.~Y. Liu, J.~Y. Zhu, A.~Tao, and B.~Catanzaro.
\newblock High-resolution image synthesis and semantic manipulation with
  conditional gans.
\newblock 2017.

\bibitem{wang2019neural}
Z.~Wang, Z.~Liu, Z.~Chen, H.~Hu, and S.~Lian.
\newblock A neural virtual anchor synthesizer based on seq2seq and gan models.
\newblock {\em arXiv preprint arXiv:1908.07262}, 2019.

\bibitem{Yang2017LearningFaceAge}
H.~Yang, H.~Di, Y.~Wang, and A.~K. Jain.
\newblock Learning face age progression: A pyramid architecture of gans.
\newblock 2017.

\bibitem{yi2017dualGAN}
Z.~Yi, H.~Zhang, P.~Tan, and M.~Gong.
\newblock Dualgan: Unsupervised dual learning for image-to-image translation.
\newblock In {\em Proceedings of the IEEE International Conference on Computer
  Vision}, pages 2849--2857, 2017.

\bibitem{yu2017seqgan}
L.~Yu, W.~Zhang, J.~Wang, and Y.~Yu.
\newblock Seqgan: Sequence generative adversarial nets with policy gradient.
\newblock In {\em Thirty-First AAAI Conference on Artificial Intelligence},
  2017.

\bibitem{zhang2018SAGAN}
H.~Zhang, I.~Goodfellow, D.~Metaxas, and A.~Odena.
\newblock Self-attention generative adversarial networks.
\newblock {\em arXiv preprint arXiv:1805.08318}, 2018.

\bibitem{zhang2017ageGAN}
Z.~Zhang, Y.~Song, and H.~Qi.
\newblock Age progression/regression by conditional adversarial autoencoder.
\newblock In {\em Proceedings of the IEEE Conference on Computer Vision and
  Pattern Recognition}, pages 5810--5818, 2017.

\bibitem{zhao2018lookAcrossElapse}
J.~Zhao, Y.~Cheng, Y.~Cheng, Y.~Yang, H.~Lan, F.~Zhao, L.~Xiong, Y.~Xu, J.~Li,
  S.~Pranata, et~al.
\newblock Look across elapse: Disentangled representation learning and
  photorealistic cross-age face synthesis for age-invariant face recognition.
\newblock {\em AAAI}, 2019.

\bibitem{zhu2017cycleGAN}
J.-Y. Zhu, T.~Park, P.~Isola, and A.~A. Efros.
\newblock Unpaired image-to-image translation using cycle-consistent
  adversarial networks.
\newblock In {\em Proceedings of the IEEE International Conference on Computer
  Vision}, pages 2223--2232, 2017.

\end{thebibliography}
}

\end{document}